\def\year{2020}\relax
\documentclass[letterpaper]{article} 
\usepackage{aaai20}  
\usepackage{times}  
\usepackage{helvet} 
\usepackage{courier}  
\usepackage[hyphens]{url}  
\usepackage{graphicx} 
\usepackage{graphicx}  
\newcommand{\citet}[1]{\citeauthor{#1}~\shortcite{#1}}
\newcommand{\citep}{\cite}

\usepackage{bm}
\usepackage[linesnumbered,ruled]{algorithm2e}
\usepackage{amsmath}
\usepackage{mathtools}
\usepackage{amssymb}
\usepackage{amsthm}
\usepackage{array}
\usepackage{color}
\usepackage{multirow}
\usepackage{comment}
\usepackage{booktabs}

\title{Fine-Tuning by Curriculum Learning for Non-Autoregressive \\
Neural Machine Translation}
\author{Junliang Guo,$^\dag$ Xu Tan,$^\ddag$ Linli Xu,$^\dag$\thanks{Corresponding Author.} Tao Qin,$^\ddag$ Enhong Chen,$^\dag$ Tie-Yan Liu$^\ddag$ \\
$^\dag$Anhui Province Key Laboratory of Big Data Analysis and Application,\\
School of Computer Science and Technology,
University of Science and Technology of China \\
$^\ddag$Microsoft Research \\
$^\dag$guojunll@mail.ustc.edu.cn, \{linlixu,cheneh\}@ustc.edu.cn,
$^\ddag$\{xuta,taoqin,tyliu\}@microsoft.com
}

\begin{document}

\maketitle

\begin{abstract}
Non-autoregressive translation (NAT) models remove the dependence on previous target tokens and generate all target tokens in parallel, resulting in significant inference speedup but at the cost of inferior translation accuracy compared to autoregressive translation (AT) models. Considering that AT models have higher accuracy and are easier to train than NAT models, and both of them share the same model configurations, a natural idea to improve the accuracy of NAT models is to transfer a well-trained AT model to an NAT model through fine-tuning. However, since AT and NAT models differ greatly in training strategy, straightforward fine-tuning does not work well. In this work, we introduce curriculum learning into fine-tuning for NAT. Specifically, we design a curriculum in the fine-tuning process to progressively switch the training from autoregressive generation to non-autoregressive generation. Experiments on four benchmark translation datasets show that the proposed method achieves good improvement (more than $1$ BLEU score) over previous NAT baselines in terms of translation accuracy, and greatly speed up (more than $10$ times) the inference process over AT baselines.

\end{abstract}

\section{Introduction}
\label{sec:intro}
\noindent Neural machine translation (NMT)~\citep{bahdanau2014neural,gehring2017convolutional,shen2018dense,vaswani2017attention,he2018layer,hassan2018achieving} has made rapid progress in recent years. The dominant approaches for NMT are based on autoregressive translation (AT), where the generation of the current token in the target sentence depends on the previously generated tokens as well as the source sentence. The conditional distribution of sentence generation in AT models can be formulated as:
\begin{equation}
\label{equ:at_manner}
P(y|x)=\prod_{t=1}^{T_{y}}P(y_{t}|y_{<t},x),
\end{equation}
where $T_{y}$ is the length of the target sentence which is implicitly decided by predicting the [\texttt{EOS}] token, and
$y_{<t}$ represents all generated target tokens before $y_{t}$ and $x$ represents the source sentence. Since AT model generates the target tokens sequentially, the inference speed is a natural bottleneck for real-world machine translation systems. 

Recently, non-autoregressive translation (NAT) models~\citep{gu2017non,kaiser2018fast,lee2018deterministic,guo2019non,wang2019non,li2019hintbased} 
are proposed to reduce the inference latency by generating all target tokens independently and simultaneously. Instead of conditioning on previously generated target tokens, NAT models generate target tokens by taking other target-independent signals as the decoder input. In this way, the generation of $y$ can be written as:
\begin{equation}
\label{equ:nat_manner}
P(y|x)=P(T_{y}|x) \cdot \prod_{t=1}^{T_{y}}P(y_{t}|z,x),
\end{equation}
where $P(T_{y}|x)$ is the explicit length prediction process for NAT models, and $z$ represents the decoder input which is generated conditionally independent of $y$.
As a result, the inference speed can be significantly boosted. However, the context dependency within the target sentence is sacrificed at the same time, which leads to a large degradation of the translation quality of NAT models.
Therefore, improving the accuracy of NAT models becomes a critical research problem.

Considering that 1) NAT is a harder task than AT due to that the decoder in the NAT model has to handle the translation task conditioned on less and weaker target-side information; 2) AT models are of higher accuracy than NAT models; 3) NAT models~\citep{gu2017non,guo2019non,wang2019non} usually share the same encoder-decoder framework with AT models~\citep{vaswani2017attention}, it is very natural to fine-tune a well-trained AT model for NAT, in order to transfer the knowledge learned in the AT model, especially the ability of target language modeling and generation in the decoder. However, AT and NAT models differ a lot in training, and thus directly fine-tuning a well-trained AT model does not lead to a good NAT model in general. 

\begin{figure*}[tb]
\centering
\centerline{\includegraphics[width=1.5\columnwidth]{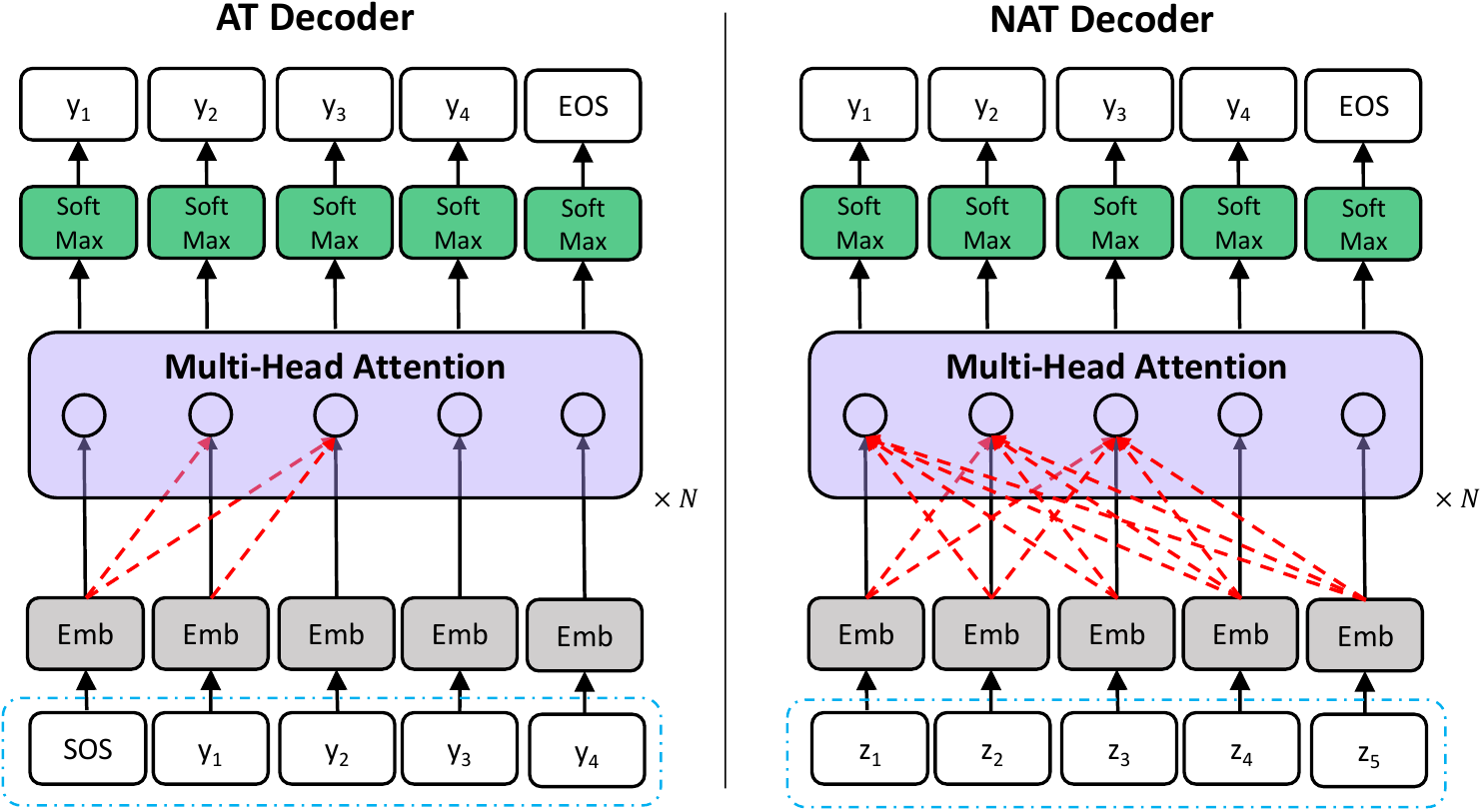}}
\caption{The comparison between the decoders of AT models and NAT models. The red dashed line indicates the attention mask, and we only draw the masks of the first three tokens for simplicity. The blue dashed box indicates the decoder input. Best view in color.}
\label{fig:compare_decoder}
\end{figure*}

To effectively transfer an AT model and obtain a good NAT model, we first note that there are two major differences between NAT and AT models, as shown in Figure~\ref{fig:compare_decoder}. 

\begin{itemize}
\item 
\noindent \textbf{Decoder input}: The decoder in AT models leverages the previous tokens as input while the decoder in NAT models takes target-independent signals as input. Specifically, \citet{gu2017non} and \citet{wang2019non} take a copy of the source sentence $x$ as the decoder input. 

\item 
\noindent \textbf{Attention mask}: Each token can only attend to the tokens in its previous positions in AT models, while each token can attend to the tokens in all positions in NAT models.
\end{itemize}

In order to handle the differences between the AT and NAT models during the fine-tuning process, we introduce the idea of curriculum learning~\cite{bengio2009curriculum} to make the transfer smooth and progressive. Specifically, 
we propose two kinds of curriculums for the transfer from an AT model to an NAT model:
\begin{itemize}
\item
\noindent \textbf{Curriculum for the decoder input}: We first feed the target sentence as AT models do, and then randomly substitute a number of tokens by the tokens in the copied source sentence, where the number of substituted tokens depends on a probability that is monotonically increasing w.r.t the training step.
\item
\noindent \textbf{Curriculum for the attention mask}: We first train the model with the attention mask of AT models and switch to that of NAT models entirely after a pre-defined training step. 
\end{itemize}

In this way, we first train the translation model in an easier autoregressive generation, and gradually transfer to a harder non-autoregressive generation. We conduct experiments on four translation datasets including WMT14 English-German, WMT14 German-English and IWSLT14 German-English to verify the effectiveness of the proposed method, and our model outperforms all non-autoregressive baselines on these tasks. Specifically, we outperform the best NAT baseline~\citep{wang2019non} by $1.87$ BLEU on the IWSLT14 De-En task and $1.14$ BLEU on the WMT14 En-De task.

\section{Related Work}

\subsection{Non-Autoregressive Neural Machine Translation}
As shown by Equation~\ref{equ:nat_manner}, NAT models generate target tokens conditioned on the source sentence $x$ and the decoder input $z$, and some previous works concentrate on the design of $z$. \citet{gu2017non} introduce a fertility predictor to guide how many times a source token is copied to the decoder input. \citet{lee2018deterministic} define $z$ by iteratively refining the target sentences generated by NAT. \citet{guo2019non} enhance the decoder input with target-side information by either utilizing auxiliary information or introducing extra parameters. \citet{ren2019fastspeech} directly use the expanded hidden sequence from the source side as $z$ in text to speech problem. Besides trying different designs of $z$, \citet{wang2019non} and \citet{li2019hintbased} propose auxiliary loss functions to solve the problem that NAT models tend to translate missing and duplicating words.

Another line of related works focuses on finding a tradeoff between high inference speed and good translation performance. Traditional AT models take $O(n)$ iterations to generate a sentence with length $n$ during inference. \citet{kaiser2018fast} 
takes intermediate discrete variables with length $m=\frac{n}{8}$ as $z$, which are generated autoregressively. These methods can result in a generation complexity of $O(\frac{n}{8})$, and similar complexity also holds for~\citet{wang2018semi} and~\citet{stern2018blockwise}. Recently, some works~\citep{stern2019insertion,welleck2019non} propose to change the generation order from the traditional left-to-right manner to a tree-based manner, resulting in a complexity of $O(\log n)$.

In this paper, we focus on NAT models with generation complexity of $O(1)$ and propose a new perspective of the training paradigm. We consider the training of AT and NAT models easier and harder tasks respectively, and utilize the strategy of fine-tuning by curriculum learning to train the model, i.e., smoothly transferring from the training of AT models to the training of NAT models.

\subsection{Transfer Learning}
Transfer learning has been extensively studied in machine learning and deep learning~\citep{pan2009survey}. 
For example, the model pre-trained on ImageNet is widely used as the initialization in downstream tasks such as object detection~\citep{girshick2014rich} and image segmentation~\citep{long2015fully}. 
On NLP tasks, the pre-trained model such as BERT~\citep{devlin2018bert} and MASS~\citep{song2019mass}
are fine-tuned in many language understanding and generation tasks.
In this paper, we find that the pre-training task (AT) and fine-tuning task (NAT) are quite different in the training strategy, where directly fine-tuning results in sub-optimal performance. Therefore, we propose using curriculum learning in the transfer process to achieve a soft landing of the AT models on NAT.

\subsection{Curriculum Learning}
Humans usually learn better when the curriculums are organized from easy to hard. Inspired by that, \citet{bengio2009curriculum} propose curriculum learning, a machine learning training strategy that feeds training instances to the model from easy to hard. Most works on curriculum learning focus on determining the order of data~\citep{lee2011learning,sachan2016easy} or tasks~\citep{pentina2015curriculum,sarafianos2017curriculum}. In our setting, we design curriculums for neither data samples nor tasks, but the training mechanisms. This way, we make the fine-tuning process smoother and ensure a soft landing from the AT models to NAT models.

\section{Fine-Tuning by Curriculum Learning for NAT}
In this section, we introduce the proposed method, Fine-tuning by Curriculum Learning for Non-Autoregressive Translation~(FCL-NAT). We start with the problem definition, and then introduce the methodology as well as some discussions on our proposed method.

\subsection{Problem Definition}
Given a source sentence $x \in \mathcal{X}$ and target sentence $y \in \mathcal{Y}$, we consider autoregressive translation~(AT) as the source task $\mathcal{T}_{S}=\{ \mathcal{Y}, P(y|z_{\textrm{AT}},x)\}$\footnote{We follow the task definition in \citet{pan2009survey}, where the tuple consists of a label space $\mathcal{Y}$ and a prediction function $P(y|z_{\textrm{AT}},x)$.}, and non-autoregressive translation~(NAT) as the target task $\mathcal{T}_{T}=\{ \mathcal{Y}, P(y|z_{\textrm{NAT}},x)\}$, where $z_{\textrm{AT}}$ and $z_{\textrm{NAT}}$ are the decoder input of AT and NAT models respectively. Given a bilingual sentence pair $(x,y)$, the conditional probability $P(y|z, x)$ can be written as
\begin{equation}
\label{equ:condition_p}
P(y|z,x)=\prod_{t=1}^{T_{y}}P(y_{t}|z,x) 
=\prod_{t=1}^{T_{y}}P(y_{t}|z,x;\theta_{\textrm{enc}}, \theta_{\textrm{dec}}),
\end{equation}
where $T_{y}$ is the length of the target sentence, $\theta_{\textrm{enc}}$ and $\theta_{\textrm{dec}}$ denote the parameters of the encoder and decoder. For AT models, we denote the decoder input as $z_{\textrm{AT}}=(y_{0}, ..., y_{t-1})$, which is the left-shifted target sentence in teacher forcing~\citep{williams1989learning} training. For NAT models, we denote the decoder input as $z_{\textrm{NAT}}=(\tilde{x}_{1}, ..., \tilde{x}_{T_{y}})$, which is obtained from copying the source sentence $x$. Note that we do not follow~\citep{gu2017non} which introduces a learnable neural network based fertility predictor to guide the copying process, but utilize a simple and efficient hard copy method which has been used in several previous works~\citep{wang2019non,guo2019non,li2019hintbased}.

Our objective is to learn an NAT model $\Theta=(\theta_{\textrm{enc}}, \theta_{\textrm{dec}})$, utilizing the knowledge learned in the source task $\mathcal{T}_{S}$ to facilitate learning in the target task $\mathcal{T}_{T}$, with a curriculum learning way to fine-tune from
\begin{equation}
\label{equ:at_loss}
L_{\textrm{AT}}(x,y;\Theta) = -\sum_{t=1}^{T_{y}} \log P(y_{t}|z_{\textrm{AT}},x)
\end{equation}
to
\begin{equation}
\label{equ:nat_loss}
L_{\textrm{NAT}}(x,y;\Theta) = -\sum_{t=1}^{T_{y}} \log P(y_{t}|z_{\textrm{NAT}},x).
\end{equation}

\subsection{Methodology}
\label{sec_3_method}
Generally, the transfer learning procedure for our NAT model
can be divided into three stages: 
1) AT training, where the model is trained autoregressively as the traditional AT model, and this is equivalent to initializing our model with a pre-trained AT model; 2) Curriculum learning, which is the main stage we focus on and will be introduced with details in this section; 3) NAT training, where we train the model non-autoregressively until convergence. We denote the training steps of the corresponding stages as
$\textrm{I}_{\textrm{AT}}$, $\textrm{I}_{\textrm{CL}}$ and $\textrm{I}_{\textrm{NAT}}$.

As introduced in Section~\ref{sec:intro}, decoder input and attention mask are the two major differences between AT and NAT models. We describe the curriculums on the two components respectively.

\subsubsection{Curriculum for the decoder input}
For the decoder input, we implement the smooth transfer from $z_{\textrm{AT}}$ to $z_{\textrm{NAT}}$ by progressively substituting the tokens in $z_{\textrm{AT}}$ with the tokens at the same positions of $z_{\textrm{NAT}}$. Specifically, at the $i$-th training step, given the AT decoder input $z_{\textrm{AT}}$ and the NAT decoder input $z_{\textrm{NAT}}$, we first calculate the substitution rate $\alpha_{i}=f_{\textrm{sub}}(i) \in [0,1]$, where $f_{\textrm{sub}}(i)$ is the substitution function which is increased monotonically from $0$ to $1$ w.r.t the training step $i$. Then, we randomly select $n_i=\lfloor \alpha_{i} \cdot T_{y} \rfloor$ tokens to substitute. We use a binary vector $P_T \in \{0,1\}^{T_{y}}$ to represent the positions of selected tokens, where $\Vert P_T \Vert_{1}=n_i$. If the $j$-th element of $P_T$ equals $1$, it indicates the $j$-th token will be substituted. Then, the decoder input $z_{i}$ after substitution can be computed formally as:
\begin{equation}
\begin{aligned}
\label{equ:sub_result}
z_{i} &= (1-P_T) \odot z_{\textrm{AT}} + P_T \odot z_{\textrm{NAT}},
\end{aligned}
\end{equation}
where $\odot$ is the element-wise multiplication.

\begin{table}[tb]
\centering
\begin{tabular}{l|c}
\toprule
Pacing Functions & Description \\
\midrule
Ladder-like& $f_{\textrm{ladder}}(i)=\frac{\left \lfloor \frac{i + 1}{K} \right \rfloor \cdot K}{\textrm{I}_{\textrm{CL}}}$ \\
\midrule
Linear& $f_{\textrm{linear}}(i)=\frac{i + 1}{\textrm{I}_{\textrm{CL}}}$ \\
\midrule
Logarithmic& $f_{\textrm{log}}(i)=\frac{\log(i + 1)}{\log(\textrm{I}_{\textrm{CL}})}$ \\
\bottomrule
\end{tabular}
\caption{The proposed different pacing functions and their definitions.}
\label{tab:sub_functions}
\end{table}

\begin{figure}[tb]
\centering
\centerline{\includegraphics[width=0.8\columnwidth]{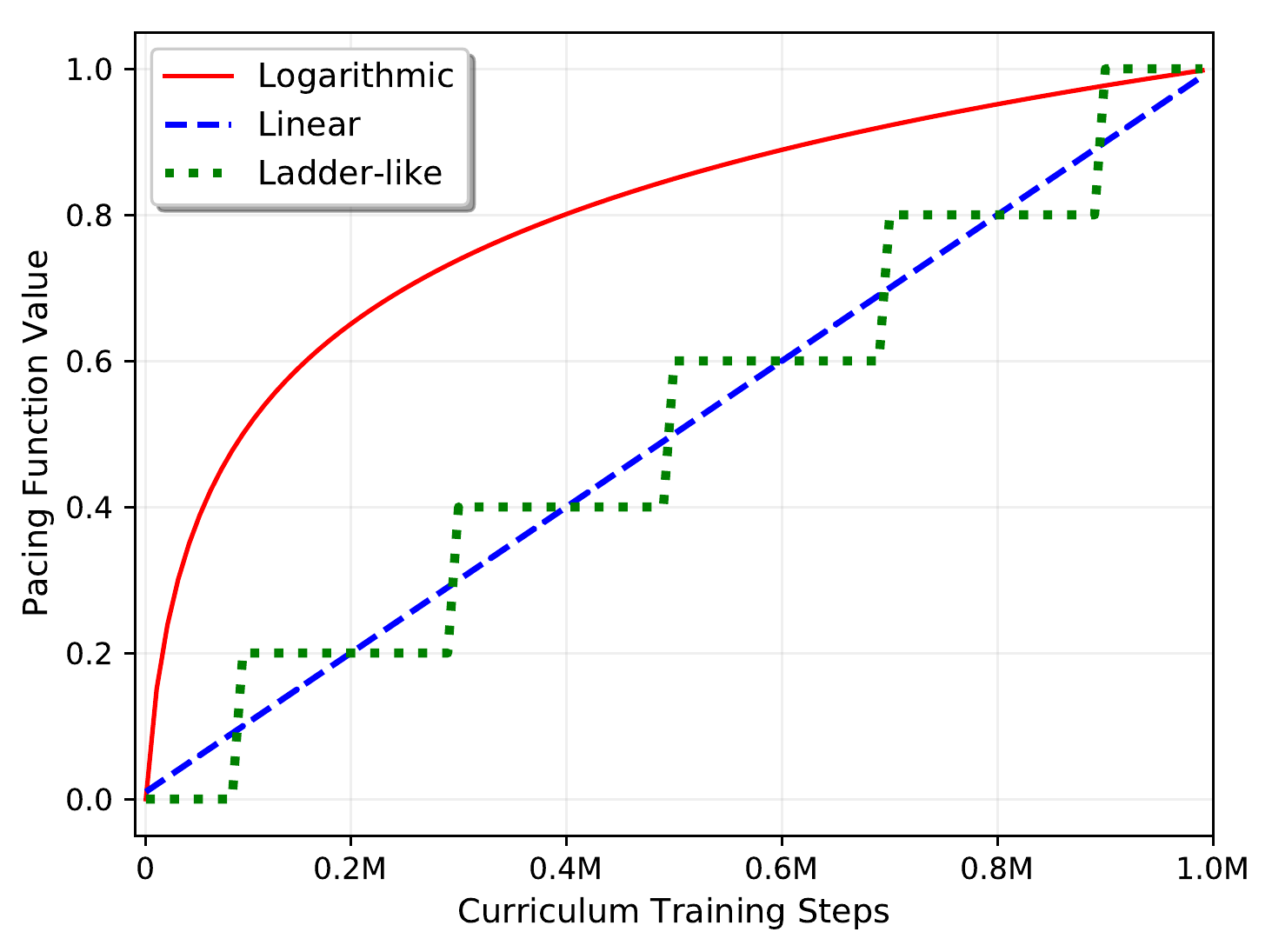}}
\caption{Illustration of the proposed pacing functions. We set $\textrm{I}_{\textrm{CL}}=1$M here, and $K=20$ for the ladder-like function, i.e., divide the training stage into $5$ sub-stages.}
\label{fig:pacing_functions_illus}
\end{figure}

The substitution function $f_{\textrm{sub}}(i)$ controls how fast the decoder input will be transfered from $z_{\textrm{AT}}$ to $z_{\textrm{NAT}}$, and we term it as the pacing function because it works similar to pacing strategies introduced in previous curriculum learning works~\citep{kumar2010self,hacohen2019power}. In this paper, we define three different pacing functions named as ladder-like, linear and logarithmic to make a smooth transformation from $z_{\textrm{AT}}$ to $z_{\textrm{NAT}}$ and verify the effectiveness of the proposed model.

The definition and illustration of these pacing functions are shown in Table~\ref{tab:sub_functions} and Figure~\ref{fig:pacing_functions_illus} respectively.
Correspondingly, the ladder-like function divides the curriculum learning stage into $\frac{\textrm{I}_{\textrm{CL}}}{K}$ sub-stages, and the substitution rate is fixed within each sub-stage but increased when switching to the next sub-stage. As this is a discrete pacing strategy w.r.t the training step $i$, we propose the other two continuous functions for comparison. According to their definitions, the difference between linear and logarithmic pacing functions is that $f_{\textrm{linear}}(i)$ does not show any preference on easier or harder stages but keeps a steady increasing pace, and $f_{\textrm{log}}(i)$ concentrates more on the harder stage. As different substitution functions reflect different pacing strategies in curriculum learning, we compare and analyze the proposed functions in experiments.

\subsubsection{Curriculum for the attention mask}

\begin{algorithm}
\caption{Fine-tuning by curriculum learning for NAT~(FCL-NAT)}
\label{alg:cl_nat}
  \SetKwInOut{Input}{Input} 
    \Input{The translation model $\Theta = (\theta_{\textrm{enc}}, \theta_{\textrm{dec}})$; the training set $(\mathcal{X}, \mathcal{Y})$; the AT/NAT decoder input $z_{\textrm{AT}}$ and $z_{\textrm{NAT}}$, the attention mask $M_{\textrm{AT}}$ and $M_{\textrm{NAT}}$; the substitution function $f_{\textrm{sub}}(i)$ which is chose from aforementioned functions; the maximum curriculum training step $\textrm{I}_{\textrm{CL}}$; the mask switching rate $\alpha_{\textrm{M}}$.}
    \BlankLine
    Set the attention mask $M=M_{\textrm{AT}}$ \;
    Pretrain the model autoregressively following the loss function Equation~\ref{equ:at_loss} \;
    \For{$i=1...\textrm{I}_{\textrm{CL}}$}{
      Draw a mini-batch of training pairs from $(\mathcal{X}, \mathcal{Y})$ \;
      Compute the substitution rate $\alpha_{i}=f_{\textrm{sub}}(i)$ and construct the substitution mask $P_T$ \;
      \uIf{$\alpha_{i} > \alpha_{\textrm{M}}$}{
        Set the attention mask $M=M_{\textrm{NAT}}$ \;
        }
      Compute the substitution result $z_{i}$ following Equation~\ref{equ:sub_result} \;
      Train the model with the loss $L_{\textrm{CL}}(x,y;\Theta) = -\sum_{t=1}^{T_{y}} \log P(y_{t}|z_{i},x)$ on this mini-batch \;
     }
    Train the model in purely non-autoregressive manner until convergence following Equation~\ref{equ:nat_loss}.
\end{algorithm}

The attention mask can be formulated as a zero-one matrix $M \in \{0,1\}^{T_{y} \times T_{y}}$. For AT models, $M_{\textrm{AT}}$ is an upper triangle matrix to prevent the model from attending to future words during training. For NAT models, as the decoder input is the copied source sentence and all target words are generated conditionally independently, and thus the matrix $M_{\textrm{NAT}}$ is a matrix with all ones. We find that there does not exist a natural intermediate state between these two types of attention masks, therefore we choose to directly switch the mask from AT to NAT when the substitution rate exceeds a pre-defined threshold $\alpha_{\textrm{M}}$.

We summarize the whole procedure of fine-tuning by curriculum learning in Algorithm~\ref{alg:cl_nat}. If we remove the curriculum learning part from line $3$ to line $10$, it yields the traditional fine-tuning strategy, which will be compared as a baseline in our experiments.

\subsection{Discussion}
\label{sec:discuss}

\noindent \textbf{Why token-level substitution?}~~~A straightforward idea of implementing the curriculum learning procedure from AT to NAT models is to conduct sentence-level substitution, i.e., we can directly replace the decoder input from $z_{\textrm{AT}}$ to $z_{\textrm{NAT}}$ at the sentence level with probability $\alpha_{i}$ at the $i$-th training step. However, the sentence-level substitution is only alternating between two different training strategies, without explicitly providing a transfer to sufficiently leverage the information contained in the intermediate states between AT and NAT models. Our preliminary experiments verify our statement by showing that the performance of sentence-level substitution is inferior to that of token-level substitution under the same setting,
and detailed results are listed in the section of experiments.

\noindent \textbf{Why directly switch the attention mask?}~~~The decoder input of AT models $z_{\textrm{AT}}$ is the left-shifted target sequence, e.g., the $(i+1)$-th token is the label of the $i$-th token, and $M_{\textrm{AT}}$ prevents the $i$-th token from seeing its label. However, the attention mask of NAT decoder enables the $i$-th position to see the tokens in all positions. Therefore, at the early stage of curriculum learning where AT tokens are dominant in the substituted results, if we follow the same token-level substitution mechanism for the attention mask, then the substituted position will see the next AT token which is supposed to be the label of the current position, making the model learn to copy instead of learning to translate. Therefore, we use the attention mask of AT models $M_{\textrm{AT}}$ in the early stage of curriculum learning, and switch to utilize $M_{\textrm{NAT}}$ when there are enough NAT tokens in the substitution results.

\section{Experiments and Results}

\subsection{Experimental Setup}
\subsubsection{Datasets}
We evaluate our method on four widely used benchmark datasets: IWSLT14 German to English translation~(IWSLT14 De-En)
and WMT14 English to German/German to English translation~(WMT14 En-De/De-En)\footnote{https://www.statmt.org/wmt14/translation-task}. We strictly follow the dataset configurations of previous works~\citep{gu2017non,guo2019non}. Specifically, for the IWSLT14 De-En task, we have $153k/7k/7k$ parallel bilingual sentences in the training/dev/test sets respectively. 
WMT14 En-De/De-En has a much larger dataset which contains $4.5$M training pairs, where \texttt{newstest2013} and \texttt{newstest2014} are used as the validation and test set respectively. For each dataset, we tokenize the sentences by Moses~\citep{koehn2007moses} and segment each word into subwords using Byte-Pair Encoding~(BPE)~\citep{sennrich2015neural}, resulting in a $32$k vocabulary shared by source and target languages.

\subsubsection{Model Configurations}
We follow~\citep{gu2017non,guo2019non} for the basic configuration of our model, which is based on the Transformer~\citep{vaswani2017attention} architecture that consists of multi-head attention and feed-forward networks. We also utilize the multi-head positional attention proposed by~\citep{gu2017non}. For WMT14 datasets, we use the hyperparameters of a \texttt{base} transformer ($d_{\textrm{model}}=d_{\textrm{hidden}}=512$, $n_{\textrm{layer}}=6$, $n_{\textrm{head}}=8$). For IWSLT14
datasets, we utilize smaller architectures ($d_{\textrm{model}}=d_{\textrm{hidden}}=256$, $n_{\textrm{layer}}=5$, $n_{\textrm{head}}=4$) for IWSLT14.
Please refer to~\citep{vaswani2017attention,gu2017non} for more detailed settings.

\subsubsection{Training and Inference}
Following previous works~\citep{gu2017non,lee2018deterministic,guo2019non}, we also utilize sequence-level knowledge distillation~\citep{kim2016sequence} during training. We first train an AT teacher model which has the same architecture as the NAT student model, then we use the translation results of each source sentence generated by the teacher model as the new ground truth to formulate a new training set. The distilled training set is more deterministic and less noisy, and thus makes the training of NAT models much easier~\citep{gu2017non}. We set the beam size to be $4$ for the teacher model. 
While the performance of the AT teacher may influence the performance of the NAT student~\citep{wang2019non}, to ensure a fair comparison,
we use the autoregressive models of the same performance with that in~\citep{wang2019non} as our teacher models for all datasets.
We train the NAT model on $8/1$ Nvidia M40 GPUs for WMT/IWSLT datasets respectively, and follow the optimizer setting in Transformer~\citep{vaswani2017attention}. 
We adopt the logarithmic pacing function for the main results. For the three training stages introduced in Section~\ref{sec_3_method}, we list their settings in Table~\ref{tab:training_step_set}, which are determined by the model performance on the validation sets.
We set $\alpha_{\textrm{M}}=0.6$ for all tasks.
We implement our model on Tensorflow\footnote{https://github.com/tensorflow/tensor2tensor},
and we have released our code\footnote{https://github.com/lemmonation/fcl-nat}.

\begin{table}[tb]
\centering
  \begin{tabular}{l|ccc}
  \toprule
  \multicolumn{1}{c|}{}& \multicolumn{2}{c}{\textbf{WMT14}} & \multicolumn{1}{c}{\textbf{IWSLT14}} \\
      & \multicolumn{1}{c}{En$-$De} & \multicolumn{1}{c}{De$-$En} & \multicolumn{1}{c}{De$-$En} \\
  \midrule
  $\textrm{I}_{\textrm{AT}}$ & $119$k & $138$k & $55$k \\
  \midrule
  $\textrm{I}_{\textrm{CL}}$ & $0.5$M & $0.5$M & $1.0$M \\
  \midrule
  $\textrm{I}_{\textrm{NAT}}$ & $1.5$M & $1.5$M & $2.0$M \\
  \bottomrule
  \end{tabular}
  \caption{Training steps for the three training stages.}
  \label{tab:training_step_set}
\end{table}

\begin{table*}[tb]
\centering
\begin{tabular}{l|ccc|rr}
\toprule
\multicolumn{1}{c|}{}& \multicolumn{2}{c}{\textbf{WMT14}}& \multicolumn{1}{c|}{\textbf{IWSLT14}} \\
\textbf{Models}   & \multicolumn{1}{c}{En$-$De} & \multicolumn{1}{c}{De$-$En} & \multicolumn{1}{c|}{De$-$En} & \multicolumn{2}{c}{Latency / Speedup}  \\
\midrule
Transformer~\citep{vaswani2017attention}    & $27.30$ & $31.29$ & $33.52$ & $607$ ms & $1.00 \times$\\
\midrule
NAT-FT~\citep{gu2017non}                 & $17.69$ & $21.47$ &  $20.32^{\dagger}$ & $39$ ms & $15.6 \times$\\
NAT-FT (NPD $10$)     & $18.66$ & $22.41$ & $21.39^{\dagger}$ & $79$ ms & $7.68 \times$\\
NAT-FT (NPD $100$)    & $19.17$ & $23.20$ & $24.21^{\dagger}$ & $257$ ms & $2.36 \times$ \\
NAT-IR~\citep{lee2018deterministic}      & $21.61$ & $25.48$ & $23.94^{\dagger}$ & $404^{\dagger}$ ms & $1.50 \times$\\
ENAT~\citep{guo2019non}  & $20.65$ & $23.23$ & $25.09$ & $24$ ms & $25.3 \times$ \\
ENAT (NPD $9$)  & $24.28$ & $26.67$ & $28.60$ & $49$ ms & $12.4 \times$ \\
NAT-Reg~\citep{wang2019non}  & $20.65$ & $24.77$ & $23.89$ & $22$ ms & $27.6 \times$ \\
NAT-Reg (NPD $9$)          & $24.61$ & $28.90$ & $28.04$ & $40$ ms & $15.1 \times$ \\
\midrule
Direct Transfer & $20.23$ & $23.16$ & $23.05$ & $21$ ms & $28.9 \times$ \\
\midrule
\textbf{FCL-NAT} & $21.70$ & $25.32$ & $26.62$ & $21$ ms & $28.9 \times$ \\
\textbf{FCL-NAT} (NPD $9$) & $\textbf{25.75}$ & $\textbf{29.50}$ & $\textbf{29.91}$ & $38$ ms & $16.0 \times$ \\
\bottomrule
\end{tabular}
\caption{The BLEU scores of our proposed FCL-NAT and the baseline methods on the WMT14 En-De, WMT14 De-En and IWSLT14 De-En tasks. ``$\dagger$'' indicates that the result is provided by~\citet{wang2019non}, and ``/'' indicates the corresponding result is not reported in the original paper. We report the best results for baseline methods and also list the inference latency as well as the speedup w.r.t autoregressive models. NPD $9$ indicates results of noisy parallel decoding with $9$ candidates, i.e., $B=4$, otherwise $B=0$.
}
\label{tab:bleu_results}
\end{table*}

During inference, we utilize Noisy Parallel Decoding~(NPD) to generate multiple samples and select the best translation from them, which is also a common practice in previous NAT models~\citep{wang2019non,guo2019non,li2019hintbased}. Specifically, as we do not know the lengths of target sentences during inference, we generate multiple translation candidates with different target lengths in $T_y \in \big[ \lfloor \beta \cdot T_{x}-B \rfloor, \lfloor \beta \cdot T_{x} + B \rfloor \big]$
where $\beta$ is the average ratio between target and source sentence lengths calculated in the training set,
and $B$ is half of the searching window of the target length. For example, $B=0$ represents greedy search. For $B \ge 1$, we first generate $2B+1$ translation candidates, and then utilize the AT teacher model to score and select the best translation as our final result. As this scoring procedure is fully parallelizable, it will not hurt the non-autoregressive property of the model. In our experiments, we set $\beta = 1.1$ for all English to German tasks, and $\beta = 0.9$ for all German to English tasks. We test with $B=0$ and $B=4$ to keep consistent with our baselines~\citep{wang2019non,guo2019non}.
We use tokenized case-sensitive BLEU~\citep{papineni2002bleu} for the WMT14 
datasets, and tokenized  case-insensitive BLEU for the IWSLT14 dataset, which are all common practices in the literature~\citep{gu2017non,wang2019non,guo2019non}. For the inference latency, we report the per-sentence decoding latency on the \texttt{newstest2014} test set of the WMT14 En-De task, i.e., set the batch size to $1$ and calculate the average translation time over all sentences in the test set, which is conducted on a single Nvidia P100 GPU to ensure a fair comparison with baselines~\citep{gu2017non,wang2019non,guo2019non}.

\subsection{Results}

We compare our model with non-autoregressive baselines including NAT with Fertility~(NAT-FT)~\citep{gu2017non}, NAT with Iterative Refinement~(NAT-IR)~\citep{lee2018deterministic}, NAT with Enhanced Decoder Input~(ENAT)~\citep{guo2019non} and NAT with Auxiliary Regularization~(NAT-Reg)~\citep{wang2019non}. For NAT-IR, we report their best results with $10$ refinement iterations. For ENAT and NAT-Reg, we report their best results when $B=0$ and $B=4$ correspondingly. We take Direct Transfer (DT) as another baseline, where we omit the curriculum learning strategy from line $3$ to line $10$ in Algorithm~\ref{alg:cl_nat}, and train the model in a non-autoregressive manner for extra $\textrm{I}_{\textrm{CL}}$ steps to ensure a fair comparison.

The main results of this paper are listed in Table~\ref{tab:bleu_results}. Our method FCL-NAT achieves significant improvements over all NAT baselines on different tasks. Specifically, note that although we do not introduce any auxiliary loss functions or new parameters, we outperform NAT-Reg and ENAT with a large margin, which demonstrates the superiority of the proposed fine-tuning by curriculum learning method. Compared with Direct Transfer, FCL-NAT brings a large improvement on translation accuracy, demonstrating the importance of the progressive transfer between two tasks with curriculum learning. As for the inference efficiency, we achieve a $16.0$ times speedup, which is comparable with NAT-Reg and ENAT.

\subsection{Analyses}

\subsubsection{Comparison with Direct Transfer}

\begin{figure}[tb]
\centering
\includegraphics[width=0.8\linewidth]{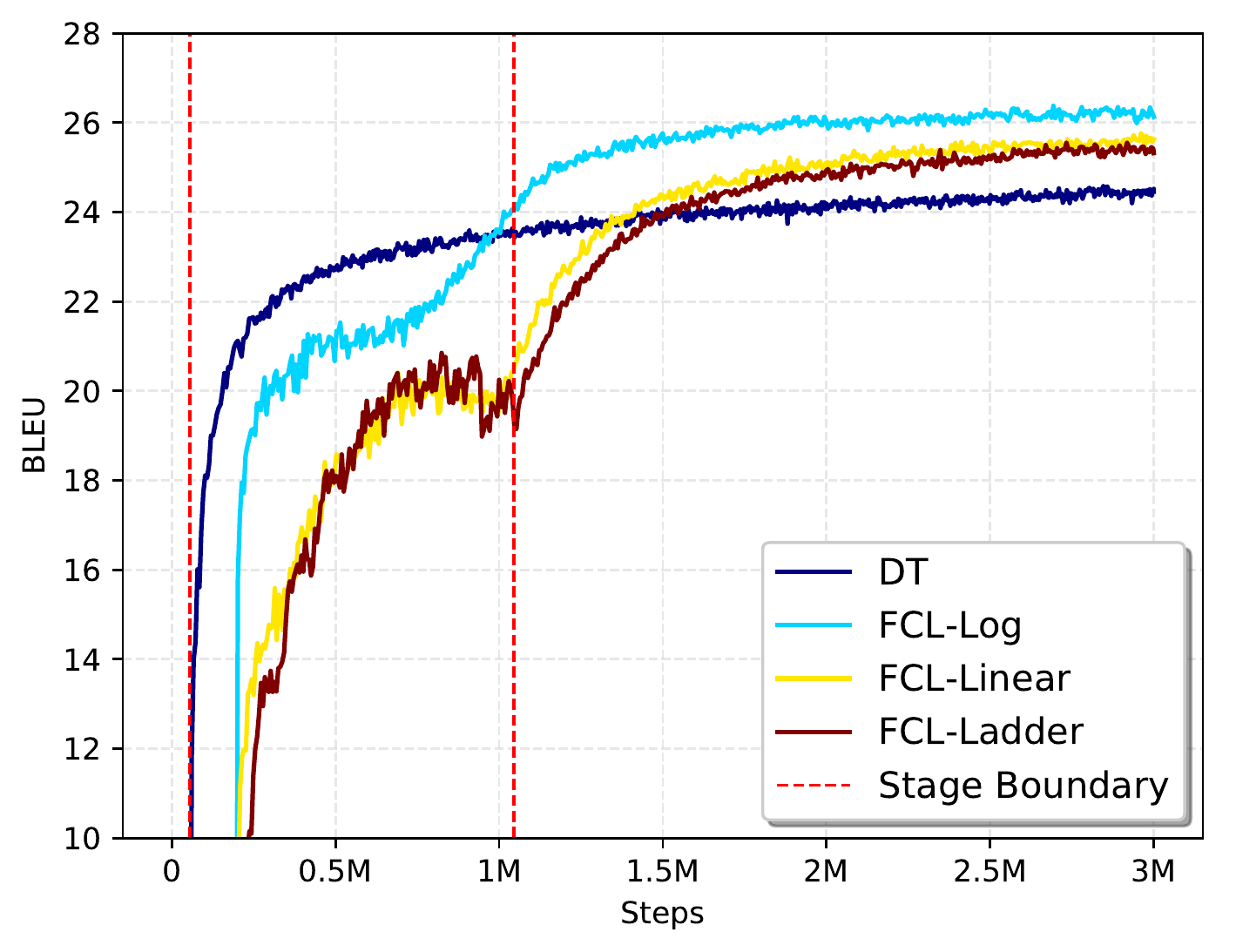}
\caption{The comparison of BLEU scores on the validation set of IWSLT14 De-En task among different pacing functions proposed in Table~\ref{tab:sub_functions} as well as the direct transfer (DT) baseline. We set $B=0$ here. The two red dashed vertical lines indicate the boundary of three training stages, i.e., AT training, curriculum learning, and NAT training from left to right.}
\label{fig:pacing_functions}
\end{figure}

We compare the training curve of our proposed FCL-NAT with Direct Transfer (DT). We evaluate FCL-NAT with different pacing functions proposed in Table~\ref{tab:sub_functions}, as well as DT on the validation set of the IWSLT14 German-to-Engish task, and plot the training curves in Figure~\ref{fig:pacing_functions}. 
We can find that the NAT models trained with different curriculum mechanisms (pacing functions) achieve better translation accuracy than the model trained with DT. 

As described in Section~\ref{sec_3_method}, the training process of the proposed method can be divided into three stages: AT training, curriculum learning and NAT training. An interesting finding from Figure~\ref{fig:pacing_functions} is that although the accuracy of our method is worse than DT in the first two stages, it finally becomes higher than DT in the NAT training stage. The worse performance in the first two stages is due to that the training in the first two stages
of our method is not consistent with the NAT inference, while there is no such issue for DT. However, our method has a higher increasing rate in the final NAT training stage and eventually outperforms DT. Clearly, the proposed transfer learning by curriculum learning strategy helps the model find a better initial point for NAT training. Due to the difference between AT and NAT models, direct initializing by AT model is far from enough for an NAT model. Our method is able to achieve a smooth transfer between AT and NAT models.

\subsubsection{Comparison of Different Pacing Functions}
The performance of different pacing functions can also be found in Figure~\ref{fig:pacing_functions}. For the ladder-like pacing function, we set $K=10$ to divide the curriculum learning stage to $10$ substages. We have two observations: 1) the ladder-like and linear pacing functions result in similar accuracy curves, and 2) the logarithmic pacing function outperforms the above two functions. These observations indicate that 1) whether the pacing function is discrete or continuous
does not influence the performance much in a constant pacing strategy, and 2) the logarithmic pacing function results in more training steps with larger substitution rates, which achieves a good trade-off between leveraging the information of AT models and training non-autoregressively,
and thus demonstrates better performance.

\subsubsection{Study on Sentence-Level Substitution}

\begin{figure}[tb]
\centering
\centerline{\includegraphics[width=0.8\columnwidth]{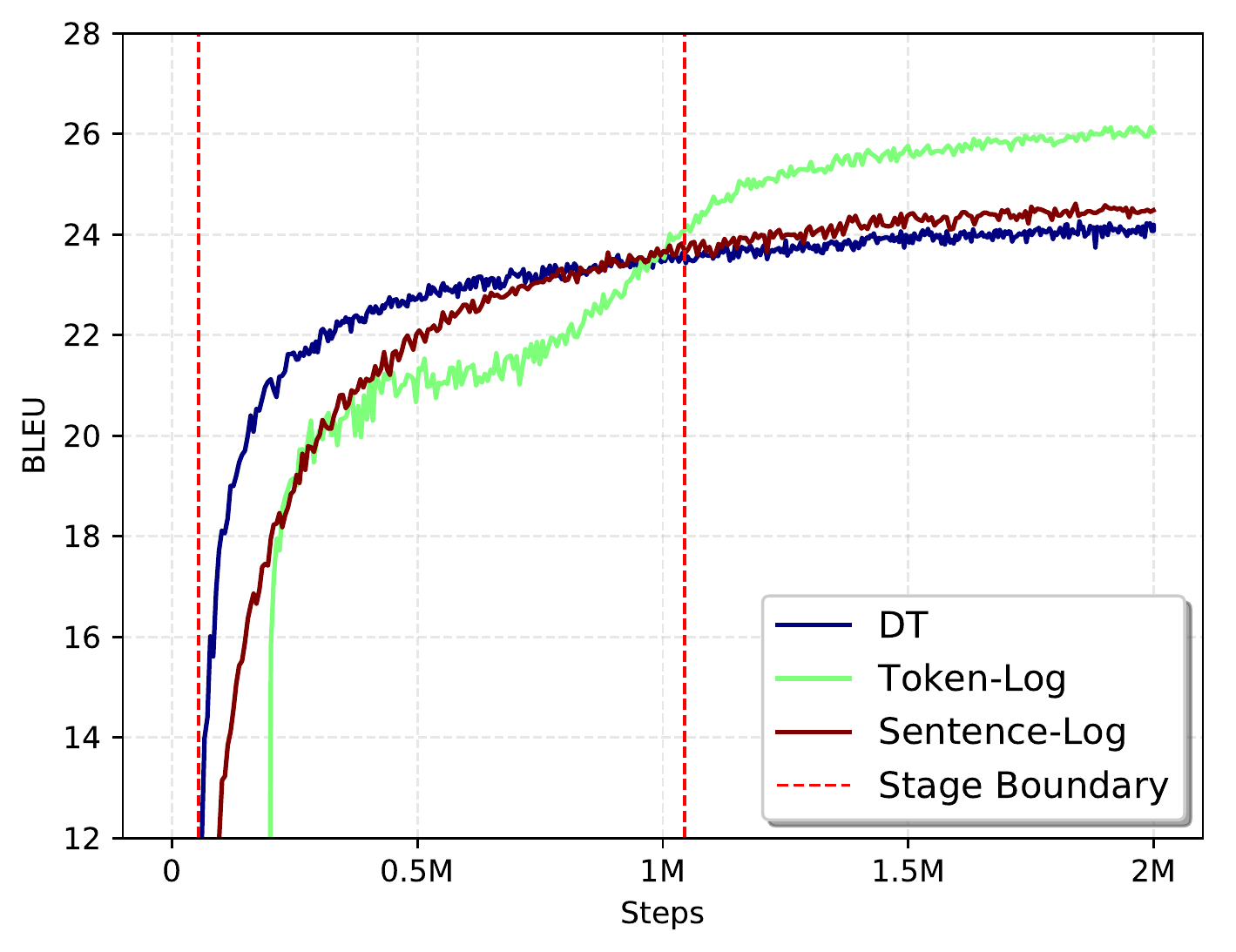}}
\caption{The comparison of BLEU scores on the validation set of the IWSLT14 De-En task between token-level and sentence-level substitution strategy as well as the direct transfer (DT) baseline. We choose the logarithmic pacing function for both substitution strategies and term them as Token-Log and Sentence-Log. The two red dashed vertical lines indicate the boundary of the three training stages, and we set $B=0$ here.}
\label{fig:compare_sen_sub}
\end{figure}

As stated in Section~\ref{sec:discuss}, a straightforward way to implement our idea is to directly replace the decoder input from $z_{\textrm{AT}}$ to $z_{\textrm{NAT}}$ when conducting substitution, termed as sentence-level substitution. We compare token-level and sentence-level substitution strategies on the validation set of the IWSLT14 De-En task, and keep other settings aligned, i.e., we choose the logarithmic pacing function and set $\textrm{I}_{\textrm{AT}}=55$k, $\textrm{I}_{\textrm{CL}}=1.0$M, $\textrm{I}_{\textrm{NAT}}=0.5$M in both settings. 
The results are shown in Figure~\ref{fig:compare_sen_sub}. 
Both token-level and sentence-level substitution strategies outperform the direct transfer baseline, which further demonstrates the efficacy of the proposed fine-tuning by curriculum learning methodology. In addition, the token-level substitution outperforms the sentence-level substitution by a large margin, showing that it is crucial to leverage the information provided by the intermediate states during transferring from a task to another.

\subsubsection{Study on Repetitive Tokens}

\begin{table}[tb]
\centering
  \begin{tabular}{c|c|c}
  \toprule
  NAT-FT & NAT-Reg & FCL-NAT \\
  \midrule
  $2.30$ & $0.90$ & $\textbf{0.57}$ \\
  \bottomrule
  \end{tabular}
  \caption{The comparison on the average number of per-sentence repetitive tokens on the validation set of the IWSLT14 De-En task.}
  \label{tab:dedup_rate}
\end{table}

As pointed out by~\citet{wang2019non}, a typical translation error of the basic NAT model is translating repetitive tokens, and they propose an auxiliary regularization function to explicitly address the problem. While our proposed FCL-NAT is not specifically designed to deal with this issue, we find it also alleviates this problem as a byproduct. We calculate the average number of consecutive repetitive tokens in a sentence on the validation set of IWSLT14 De-En, and the results are listed in Table~\ref{tab:dedup_rate}. We observe that without an explicit regularization, our model is still able to reduce repetitive tokens more effectively.

\section{Conclusion}
In this paper, we propose a novel fine-tuning by curriculum learning method for non-autoregressive neural machine translation, which progressively transfers the knowledge learned in AT models into NAT models. We consider AT training as a source and easier task and NAT training as a target and harder task, and designed a curriculum to gradually substitute the decoder input and attention mask in an AT decoder with that in an NAT decoder. Experiments on four benchmark datasets demonstrate the effectiveness of our proposed method for non-autoregressive translation. 

In the future, we will extend our idea to other tasks such as text-to-speech and image-to-image translation. As long as there exists a smooth transformation between the source task and the target task, similar ideas can be applied to leverage the intermediate states between the two tasks. In addition, it is also interesting to explore the theoretical explanation of the proposed model.

\section*{Acknowledgements}
This research was supported by the National Natural Science Foundation of China (No. 61673364, No. U1605251) and the Fundamental Research Funds for the Central Universities (WK2150110008).
The authors would like to thank Information Science Laboratory Center of USTC for the hardware and software services.

\bibliography{AAAI-GuoJ.2729}
\bibliographystyle{aaai}
\end{document}